\documentclass[11pt,a4paper]{article}
\usepackage[hyperref]{acl2019}
\usepackage{times}
\usepackage{latexsym}

\usepackage{url}

\usepackage{amssymb}
\usepackage{amsmath}
\usepackage{arydshln}

\usepackage{forest}

\usepackage{algorithm}
\usepackage[noend]{algpseudocode}
\algnewcommand{\IfThen}[2]{\State \algorithmicif\ #1\ \algorithmicthen\ #2}

\usepackage{tikz}
\usepackage{tikz-qtree}
\tikzset{every tree node/.style={align=center, anchor=north}}
\usetikzlibrary{shapes, shapes.geometric, arrows, cd, positioning, automata}
\tikzstyle{boxneuron} = [rectangle, minimum width=1.5cm, minimum height=1cm, text centered, draw=black]
\tikzstyle{arrow} = [thick,->,>=stealth]

\newcommand{\expec}{\mathop{\mathbb{E}}}
\DeclareMathOperator{\Relu}{ReLU}
\DeclareMathOperator{\softmax}{softmax}
\DeclareMathOperator*{\argmax}{argmax}
\DeclareMathOperator{\tree}{Tree}

\aclfinalcopy 

\title{Finding Syntactic Representations in Neural Stacks}

\author{\textbf{William Merrill\thanks{\; Work completed while the author was at Yale University.}\:\:\footnotemark[2]\: \hspace{2em} Lenny Khazan\footnotemark[2]\: \hspace{2em} Noah Amsel\footnotemark[2]\:} \\ \textbf{Yiding Hao\footnotemark[2]\:  \hspace{2em} Simon Mendelsohn\footnotemark[2]\:  \hspace{2em}  Robert Frank\footnotemark[2]\:} \\
\footnotemark[2]\: Yale University, New Haven, CT, USA\\
\footnotemark[1]\: Allen Institute for Artificial Intelligence, Seattle, WA, USA \\
	{\tt first.last@yale.edu}\\}
	
\date{}

\begin{document}
\maketitle
\begin{abstract}
    Neural network architectures have been augmented with differentiable stacks in order to introduce a bias toward learning hierarchy-sensitive regularities. It has, however, proven difficult to assess the degree to which such a bias is effective, as the operation of the differentiable stack is not always interpretable.  In this paper, we attempt to detect the presence of latent representations of hierarchical structure through an exploration of the unsupervised learning of constituency structure. Using a technique due to \citet{shenNeuralLanguageModeling2018,shenStraightTreeConstituency2018}, we extract syntactic trees from the pushing behavior of stack RNNs trained on language modeling and classification objectives. We find that our models produce parses that reflect natural language syntactic constituencies, demonstrating that stack RNNs do indeed  infer linguistically relevant hierarchical structure.
\end{abstract}

\section{Introduction}

Sequential models such as long short-term memory networks (LSTMs; \citealp{hochreiterLongShortTermMemory1997}) have been proven capable of exhibiting qualitative behavior that reflects a sensitivity to regularities that are structurally conditioned, such as subject--verb agreement \citep{linzenAssessingAbilityLSTMs2016,gulordavaColorlessGreenRecurrent2018}. However, detailed analysis of such models has shown that this apparent sensitivity to structure does not always generalize to inputs with a high degree of syntactic complexity \citep{marvinTargetedSyntacticEvaluation2018}. These observations suggest that sequential models may not in fact be representing sentences in the kind of hierarchically organized representations that we might expect.  

Stack-structured recurrent memory units (\citealp{joulinInferringAlgorithmicPatterns2015,grefenstetteLearningTransduceUnbounded2015,yogatamaMemoryArchitecturesRecurrent2018}; and others) offer a possible method for explicitly biasing neural networks to construct hierarchical representations and make use of them in their computation. Since syntactic structures can often be modeled in a context-free manner \citep{chomskyThreeModelsDescription1956,chomskySyntacticStructures1957}, the correspondence between pushdown automata and context-free grammars \citep{chomskyContextfreeGrammarsPushdown1962} makes stacks a natural data structure for the computation of hierarchical relations. Recently, \citet{haoContextFreeTransductionsNeural2018b} have shown that stack-augmented RNNs (henceforth \textit{stack RNNs}) have the ability to learn classical stack-based algorithms for computing context-free transductions such as string reversal. However, they also find that such algorithms can be difficult for stack RNNs to learn. For many context-free tasks such as parenthesis matching, the stack RNN models they consider instead learn heuristic ``push-only" strategies that essentially reduce the stack to unstructured recurrent memory.  Thus, even if stacks allow  hierarchical regularities to be expressed, the bias that stack RNNs introduce does not guarantee that the networks will detect them.

The current paper aims to move beyond the work of \citet{haoContextFreeTransductionsNeural2018b} in two ways. While that work was based on artificially generated formal languages, this paper considers the ability of stack RNNs to succeed on tasks over natural language data.  Specifically, we train such networks on two objectives: language modeling and the \textit{number prediction task}, a classification task proposed by \citet{linzenAssessingAbilityLSTMs2016} to determine whether or not a model can capture structure-sensitive grammatical dependencies. Further, in addition to using visualizations of the pushing and popping actions of the stack RNN to assess its hierarchical sensitivity,
we use a technique proposed by  \citet{shenNeuralLanguageModeling2018,shenStraightTreeConstituency2018} to assess the presence of implicitly-represented hierarchically-organized structure through the task of unsupervised parsing. We extract syntactic constituency trees from our models and find that they produce parses that broadly reflect phrasal groupings of words in the input sentences, suggesting that our models utilize the stack in a way that reflects the syntactic structures of input sentences.

This paper is organized as follows.  \autoref{sec:network_architecture} introduces the architecture of our stack models, which extends the architecture of \citet{grefenstetteLearningTransduceUnbounded2015} by allowing multiple items to be pushed to, popped from, or read from the stack at each computational step. \autoref{sec:experimental_procedure} then describes our training procedure and reports results on language modeling and agreement classification. \autoref{sec:interpreting_stack_usage} investigates the behavior of the stack RNNs trained on these tasks by visualizing their pushing behavior. Building on this, \autoref{sec:structure_inference} describes how we adapt \citeauthor{shenNeuralLanguageModeling2018}'s (\citeyear{shenNeuralLanguageModeling2018, shenStraightTreeConstituency2018}) unsupervised parsing algorithm to stack RNNs and evaluates the degree to which the resulting parses reveal structural representations in stack RNNs. \autoref{sec:discussion} discusses our observations, and \autoref{sec:conclusion} concludes.

\section{Network Architecture} \label{sec:network_architecture}

In a stack RNN \citep{grefenstetteLearningTransduceUnbounded2015, haoContextFreeTransductionsNeural2018b}, a neural network adhering to a standard recurrent architecture, known as a \textit{controller}, is enhanced with a non-parameterized \textit{stack}. At each time step, the controller network receives an input vector $\mathbf{x}_t$ and a recurrent state vector $\mathbf{h}_{t - 1}$ provided by the controller architecture, along with a \textit{read vector} $\mathbf{r}_{t - 1}$ summarizing the top elements on the stack. The controller interfaces with the stack by computing continuous values that serve as instructions for how the stack should be modified. These instructions consist of $\mathbf{v}_t$, a vector that is pushed to the top of the stack; $d_t$, a number representing the \textit{strength} of the newly pushed vector $\mathbf{v}_t$; $u_t$, the number of items to pop from the stack; and $r_t$, the number of items to read from the top of the stack. The instructions $\langle \mathbf{v}_t, u_t, d_t, r_t \rangle$ are produced by the controller as output and presented to the stack. The stack then computes the next read vector $\mathbf{r}_t$, which is given to the controller at the next time step. This general architecture is portrayed in Figure~1. In the next two subsections, we describe how the stack computes $\mathbf{r}_t$ using the instructions $\langle \mathbf{v}_t, u_t, d_t, r_t \rangle$ and how the controller computes the stack instructions.

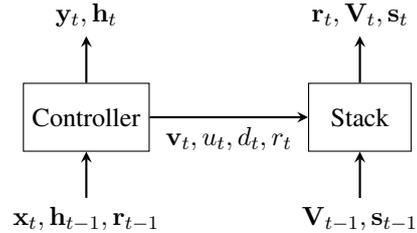
\begin{figure}
    \label{fig:stack_architecture}
	\begin{center}
		\begin{tikzpicture}[scale=1.0, every node/.style={scale=.9}]
		
		\node (controller) [boxneuron] {Controller};
		\node (stack) [boxneuron, right of=controller, xshift=3cm] {Stack};
		\node (prevstack) [below of=stack, yshift=-.5cm] {$\mathbf{V}_{t - 1}, \mathbf{s}_{t - 1}$};
		
		\node (input) [below of=controller, yshift=-.5cm] {$\mathbf{x}_t, \mathbf{h}_{t - 1}, \mathbf{r}_{t - 1}$};
		\node (output) [above of=controller, yshift=.5cm] {$\mathbf{y}_t, \mathbf{h}_t$};
		
		\node (read) [above of=stack, yshift=.5cm] {$\mathbf{r}_t, \mathbf{V}_t, \mathbf{s}_t$};
		
		\coordinate [right of=controller, xshift=1.75cm] (n1);

		\draw [arrow] (controller) -- node[below] {$\mathbf{v}_t, u_t, d_t, r_t$} (stack);
		\draw [arrow] (input) -- (controller);
		\draw [arrow] (controller) -- (output);
		\draw [arrow] (stack) -- (read);
		\draw [arrow] (prevstack) -- (stack);
		
		\end{tikzpicture}
		
		\caption{The neural stack architecture.}
	\end{center}
\end{figure}
	
\subsection{Stack Actions} \label{sec:stack_actions}

A stack at time $t$ consists of a sequence of vectors $\langle \mathbf{V}_t[1], \mathbf{V}_t[2], \dots, \mathbf{V}_t[t] \rangle$, organized into a matrix $\mathbf{V}_t$ whose $i$th row is $\mathbf{V}_t[i]$. By convention, $\mathbf{V}_t[t]$ is the ``top'' element of the stack, while $\mathbf{V}_t[1]$ is the ``bottom'' element. Each element $\mathbf{V}_t[i]$ of the stack is associated with a \textit{strength} $\mathbf{s}_t[i] \geq 0$. The strength of a vector $\mathbf{V}_t[i]$ represents the degree to which the vector is on the stack: a strength of $1$ means that the vector is ``fully'' on the stack, while a strength of $0$ means that the vector has been popped from the stack. The strengths are organized into a vector $\mathbf{s}_t = \langle \mathbf{s}_t[1], \mathbf{s}_t[2], \dots, \mathbf{s}_t[t] \rangle$. At time $t$, the stack receives a set of instructions $\langle \mathbf{v}_t, u_t, d_t, r_t \rangle$ and performs three operations: \textit{popping}, \textit{pushing}, and \textit{reading}, in that order. 

The popping operation is implemented by reducing the strength of each item on the stack by a number $\mathbf{u}_t[i]$, ensuring that the strength of each item can never fall below $0$. 
\begin{equation*}
	\mathbf{s}_t[i] = \Relu\left(\mathbf{s}_{t - 1}[i] - \mathbf{u}_t[i] \right)
\end{equation*}
The $\mathbf{u}_t[i]$s are computed as follows. The total amount of strength to be reduced is the \textit{pop strength} $u_t$. Popping begins by attempting to reduce the strength $\mathbf{s}_t[t - 1]$ of the top item on the stack by the full pop strength $u_t$. Thus, as shown below, $\mathbf{u}_t[t - 1] = u_t$. For each $i$, if $\mathbf{s}_{t - 1}[i] < \mathbf{u}_t[i]$, then the $i$th item has been fully popped from the stack, ``consuming'' a portion of the pop strength of magnitude $\mathbf{s}_{t - 1}[i]$. The strength of the next item is then reduced by an amount $\mathbf{u}_t[i - 1]$ given by the ``remaining'' pop strength $\mathbf{u}_t[i] - \mathbf{s}_{t - 1}[i]$.
\begin{align*}
	&\mathbf{u}_t[i] = \\
	&\begin{cases}
	u_t, & i = t - 1 \\
	\Relu(\mathbf{u}_t[i + 1] - \mathbf{s}_{t - 1}[i + 1]), & i < t - 1
	\end{cases}
	\end{align*}
	
The pushing operation simply places the vector $\mathbf{v}_t$ at the top of the stack with strength $d_t$. Thus, $\mathbf{V}_t$ and $\mathbf{s}_t[t]$ are updated as follows.
	\[
    \mathbf{s}_t[t] = d_t
    \qquad
    \mathbf{V}_t[i] = \begin{cases}
	\mathbf{v}_t, & i = t \\
	\mathbf{V}_{t - 1}[i], & i < t
	\end{cases} 
	\]
Note that $\mathbf{s}_t[1]$, $\mathbf{s}_t[2]$, \dots, $\mathbf{s}_t[t - 1]$ have already been updated during the popping step.

Finally, the reading operation produces a ``summary'' of the top of the stack by computing a weighted sum of all the vectors on the stack.
\[
	\mathbf{r}_t = \sum_{i = 1}^t \min\left(\mathbf{s}_t[i], \rho_t[i] \right) \cdot \mathbf{V}_t[i]
\]
The weights $\rho_t[i]$ are computed in a manner similar to the $\mathbf{u}_t[i]$s. The sum should include the top elements of the stack whose strengths add up to the \textit{read strength} $r_t$. The weight $\rho_t[t]$ assigned to the top item is initialized to the full read strength $r_t$, while the weights $\rho_t[i]$ assigned to lower items are based on the ``remaining'' read strength $\rho_t[i + 1] - \mathbf{s}_t[i + 1]$ after strength has been assigned to higher items.
\[
	\rho_t[i] = \begin{cases}
	r_t, & i = t \\
	\Relu\left( \rho_t[i + 1] - \mathbf{s}_t[i + 1] \right) & i < t
	\end{cases}
\]

\subsection{Stack Interface} \label{sec:stack_interface}

The architecture of \citet{grefenstetteLearningTransduceUnbounded2015} assumes that the controller is a neural network of the form
\begin{equation*}
    \langle \mathbf{o}_t, \mathbf{h}_t \rangle = C(\mathbf{x}_t, \mathbf{h}_{t-1}, \mathbf{r}_{t-1})
\end{equation*}
where $\mathbf{h}_t$ is its state at time $t$, $\mathbf{x}_t$ is its input, $\mathbf{r}_{t}$ is the vector read from the stack at the previous step, and $\mathbf{o}_t$ is an output vector used to produce the network output $\mathbf{y}_t$ and the stack instructions $\langle \mathbf{v}_t, u_t, d_t, r_t \rangle$.
 
The stack instructions $\langle \mathbf{v}_t, \allowbreak u_t, \allowbreak d_t, \allowbreak r_t \rangle$ are computed as follows. The read strength $r_t$ is fixed to $1$. The other values are determined by passing $\mathbf{o}_t$ to specialized layers. The vectors $\mathbf{y}_t$ and $\mathbf{v}_t$ are computed using a $\tanh$ layer, while the scalar values $u_t$ and $d_t$ are obtained from a sigmoid layer. Thus, the push and pop strengths are constrained to values between $0$ and $1$.
\begin{align}
    \mathbf{y}_t &= \softmax\left( \mathbf{W}^y\mathbf{o}_t + \mathbf{b}^y \right) \nonumber \\
    \mathbf{v}_t &= \tanh\left( \mathbf{W}^v\mathbf{o}_t + \mathbf{b}^v \right) \nonumber \\
    u_t &= \sigma\left( \mathbf{W}^u \mathbf{o}_t + b_u \right) \nonumber \\
    d_t &= \sigma\left( \mathbf{W}^d \mathbf{o}_t + b_d \right) \label{eqn:dt} \\
    r_t &= 1 \nonumber
\end{align}

This paper departs from \citeauthor{grefenstetteLearningTransduceUnbounded2015}'s architecture by allowing for push, pop, and read operations to be executed with variable strength greater than $1$. We achieve this by using an enhanced control interface inspired by \citeauthor{yogatamaMemoryArchitecturesRecurrent2018}'s (\citeyear{yogatamaMemoryArchitecturesRecurrent2018}) Multipop Adaptive Computation Stack architecture. In that model, the controller determines how much weight to pop from the stack at each time step by computing a distribution $\mathbb{P}[u]$ describing the probability of popping $u$ units from the stack. The next stack state $\mathbf{V}$ is computed as a superposition of the possible stack states $\mathbf{V}^u$ resulting from popping $u$ units from the stack, weighted by $\mathbb{P}[u]$. Our model follows \citeauthor{yogatamaMemoryArchitecturesRecurrent2018} in computing probability distributions over possible values of $u_t$, $d_t$, and $r_t$. However, instead of superimposing stack states, which may hinder interpretability, we simply set the value of each instruction to be the expected value of its associated distribution. For a distribution vector $\mathbf{p}$, define the operator $\mathbb{E}[\mathbf{p}]$ as follows:
\[
\mathbb{E}[\mathbf{p}] = \sum_{i = 0}^k i\mathbf{p}[i + 1]
\]
$\mathbb{E}[\mathbf{p}]$ denotes the expected value of $\mathbf{p}$ if we treat it as a distribution over $\lbrace 0, 1, \dots, k \rbrace$. The maximum value $k$ is fixed in advance as a hyperparameter of our model. The output $\mathbf y_t$ and instructions $\mathbf v_t$, $u_t$, $d_t$, and $r_t$ are then computed as follows:
\begin{align*}
    \mathbf{y}_t &= \softmax\left( \mathbf{W}^y\mathbf{o}_t + \mathbf{b}^y \right) \\
    \mathbf{v}_t &= \tanh\left( \mathbf{W}^v\mathbf{o}_t + \mathbf{b}^v \right) \\
    u_t &= \mathbb{E}\left[ \softmax\left( \mathbf{W}^u \mathbf{o}_t + b^u \right) \right] \\
    d_t &= \mathbb{E}\left[ \softmax\left( \mathbf{W}^d \mathbf{o}_t + b^u \right) \right] \\
    r_t &= \mathbb{E}\left[ \softmax\left( \mathbf{W}^r \mathbf{o}_t + b^r \right) \right]
\end{align*}

The full architecture that we used for language modeling and agreement classification is a controller network which, at time $t$, reads the word $\mathbf x_t$ as well as the previous stack summary $\mathbf r_{t-1}$. These vectors are passed through an LSTM layer to produce the vector $\mathbf o_t$. Then, instructions for the stack are computed from $\mathbf o_t$ according to the equations above. Finally, these instructions are executed to modify the stack state and produce the next stack summary vector $\mathbf r_t$. In our experiments, the size of the LSTM layer was 100, and the size of each stack vector was 16.

\section{Model Training} \label{sec:experimental_procedure}


This paper considers models trained on a language modeling objective and a classification objective. On each objective, we train several neural stack models along with an LSTM baseline.\footnote{Our code is available at \url{https://github.com/viking-sudo-rm/industrial-stacknns}.} This section describes the procedure used to train our models and presents the perplexity and classification values they attain on their training objectives.

\subsection{Data and Training} \label{sec:training}


Our models are trained using the \textit{Wikipedia corpus}, a subset of the English Wikipedia used by \citet{linzenAssessingAbilityLSTMs2016} for their experiments. The classification task we consider is the \textit{number prediction task}, proposed by \citet{linzenAssessingAbilityLSTMs2016} as a diagnostic for assessing whether or not LSTMs can infer grammatical dependencies sensitive to syntactic structure. In this task, the network is shown a sequence of words forming the beginning of a sentence from the Wikipedia corpus. The next word in the sentence is always a verb, and the network must predict whether the verb is singular (\textsc{sg}) or plural (\textsc{pl}). For example, on input \textit{The cats on the boat}, the network must predict \textsc{pl} to match \textit{cats}. We train and evaluate our models on the number prediction task using \citeauthor{linzenAssessingAbilityLSTMs2016}'s (\citeyear{linzenAssessingAbilityLSTMs2016}) \textit{simple dependency dataset}, which contains 141,948 training examples, 15,772 validation examples, and 1,419,491 testing examples. 

We used a model with very few parameters and basic setting of hyperparameters. The LSTM hidden state was fixed to a size of 100, while the vectors placed on the stack had size 16. Including the embedding layer, the Wikipedia model had 1,584,255 parameters. We used the Adam optimizer \citep{kingmaAdamMethodStochastic2015} with a learning rate of 0.001. The language models were trained for five epochs, while the agreement classifiers used an early stopping criterion. In addition to the LSTM baseline, for each task, we trained a stack RNN in which $u_t$ is fixed to $1$ and $d_t$ ranges from $0$ to $k = 4$, as well as a stack RNN in which $d_t$ fixed to $1$ and $u_t$ ranges from $0$ to $k = 4$. Additionally, for the classification task we trained a stack RNN in which $u_t$ ranges from $0$ to $k = 4$ and $d_t$ is computed as in \autoref{eqn:dt}.

\subsection{Evaluation} \label{sec:evaluation}


Our language models are evaluated according to two metrics. Firstly, we reserve 10\% of the Wikipedia corpus for evaluating test perplexity of the trained language models. Secondly, as a simple diagnostic of sensitivity to syntactic structure, we evaluate the performance of our Wikipedia-trained language models on \textit{number agreement prediction} \citep{linzenAssessingAbilityLSTMs2016}. Under this evaluation regime, we use our language model to simulate the number prediction task and compute the resulting classification accuracy. We do this by presenting the model with an input for the number prediction task and comparing the probabilities assigned to the verb that follows the input in the Wikipedia corpus. For example, if \textit{The cats on the boat purr} appears in the Wikipedia corpus, then we present \textit{The cats on the boat} to the language model and compare the probabilities assigned to the singular and plural forms \textit{purrs} and \textit{purr}, respectively. We consider the language model to make a correct prediction if the form of the next lexical item with the correct grammatical number (\textsc{sg} or \textsc{pl}) is predicted with greater probability than the alternative.

The number prediction classifiers we trained are evaluated according to classification accuracy. For each input sentence, we define the \textit{attractors} of the input to be the nouns intervening between the subject and the verb whose number is being classified. For example, in the input \textit{The cat on the boat}, \textit{cat} is the subject of the following verb, while \textit{boat} is an attractor. We compute the accuracy of our classifiers on the full testing set of the simple dependency data set as well as subsets of the testing set consisting of sentences with a fixed number of attractors. 

\subsection{Training Results}
\label{sec:trainig_results}

\begin{table}
    \small
	\centering
	
	\begin{tabular}{| c | c c | c | }
	    \hline
	    & \textbf{Stack} & \textbf{Stack} & \textbf{LSTM} \\
	    & ($u_t = 1$) & ($d_t = 1$) &  \\\hline
	    Perp & 92.81 & 128.28 & \textbf{91.69}  \\
	    Agree & 93.59 & 92.28 & \textbf{93.95}  \\\hline
	\end{tabular}
	\caption{Results for language models trained on the Wikipedia dataset.}
	\label{tab:lm_results}
\end{table}

\begin{table}
    \small
	\centering
	\begin{tabular}{| c | c  c c | c |}
	     \hline
	     \textbf{Number of} & \textbf{Stack} & \textbf{Stack} & \textbf{Stack} & \textbf{LSTM} \\
	     \textbf{Attractors} & ($u_t = 1$) & ($d_t = 1$) & & 
	     \\\hline
	     Overall & \textbf{98.89} & 98.88 & 98.88 & \textbf{98.89} \\
	     0 & \textbf{99.29} & 99.23 & 99.24 & 99.26 \\
	     1 & 94.75 & \textbf{95.43} & 95.18 & 95.27 \\
	     2 & 89.85 & 91.70 & \textbf{91.86} & 90.15 \\
	     3 & 83.42 & 86.59 & \textbf{87.47} & 84.30 \\
	     4 & 79.50 & 84.14 & \textbf{85.56} & 78.61 \\
	     5 & 71.07 & 71.70 & \textbf{77.99} & 74.21 \\\hline
	\end{tabular}
	
	\caption{Number prediction accuracies attained by the three stack RNN classifiers and the LSTM baseline.}
	\label{tab:classification_results}
\end{table}

\autoref{tab:lm_results} shows the quantitative results for our language models. The stack RNN is comparable to our LSTM baseline in terms of language modeling perplexity and agreement prediction accuracy when $u_t$ is fixed to $1$, though the latter performs slightly better according to both metrics. The stack RNN attains a significantly worse perplexity when $d_t$ is fixed to $1$, and its agreement prediction accuracy is worse than both the LSTM baseline and the stack RNN with $u_t = 1$.

\autoref{tab:classification_results} shows test accuracies attained by classifiers trained on the number prediction task. While the stack classifier with $u_t$ fixed to $1$ and the LSTM baseline achieve the best overall accuracy, the stack with unrestricted $u_t$ and sigmoid $d_t$ and the stack with $d_t$ fixed to $1$ exceed the baseline on sentences with at least 2 attractors. We take this to suggest that the hierarchical bias provided by the stack can improve performance on syntactically complex cases.

\section{Interpreting Stack Usage}
\label{sec:interpreting_stack_usage}

\begin{figure}
        \centering
	     \includegraphics[scale=0.32]{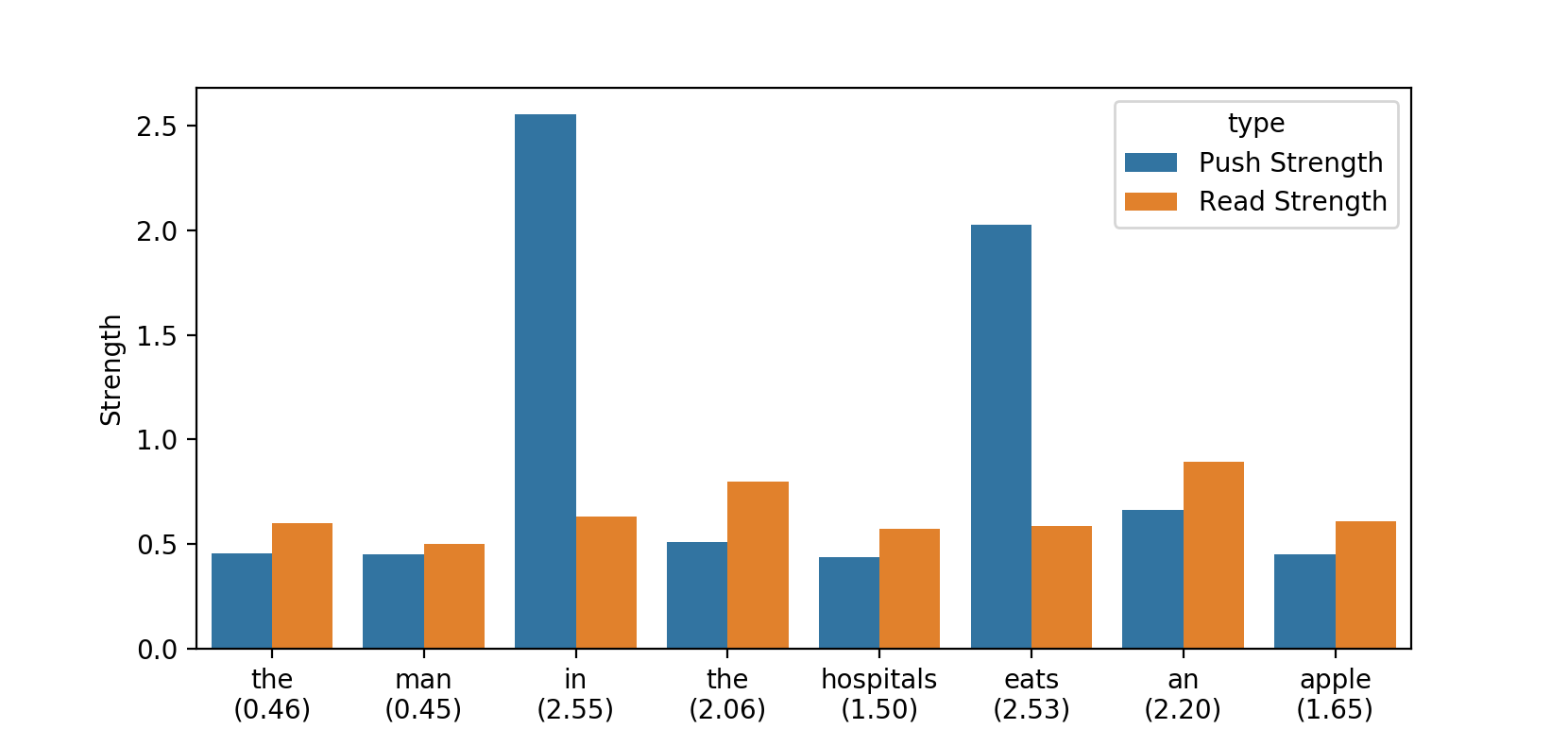}
	     \includegraphics[scale=0.32]{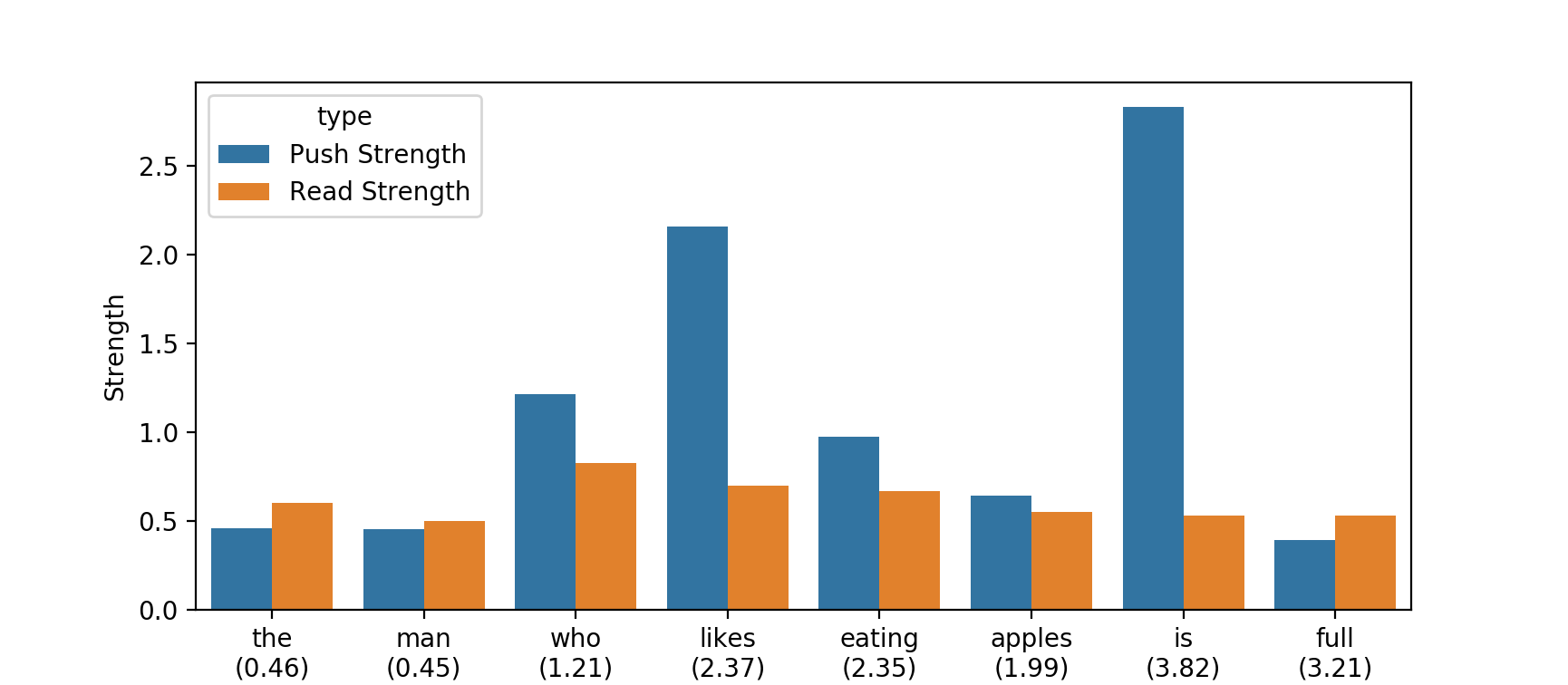} 
	    \caption{Push and read strengths computed by the $u_t = 1$ language model. Values underneath each word show the total strength remaining on the stack at that step.}
	    \label{fig:sent1_trace}
\end{figure}
\begin{figure*}[t]
	    \centering
	     \includegraphics[scale=0.6]{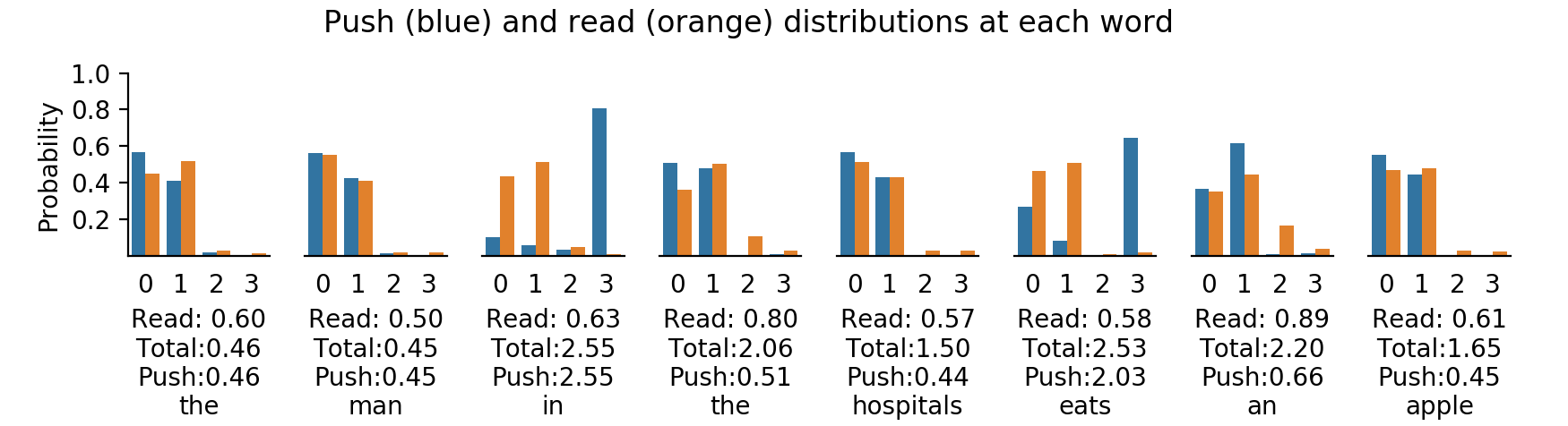}
	     \includegraphics[scale=0.6]{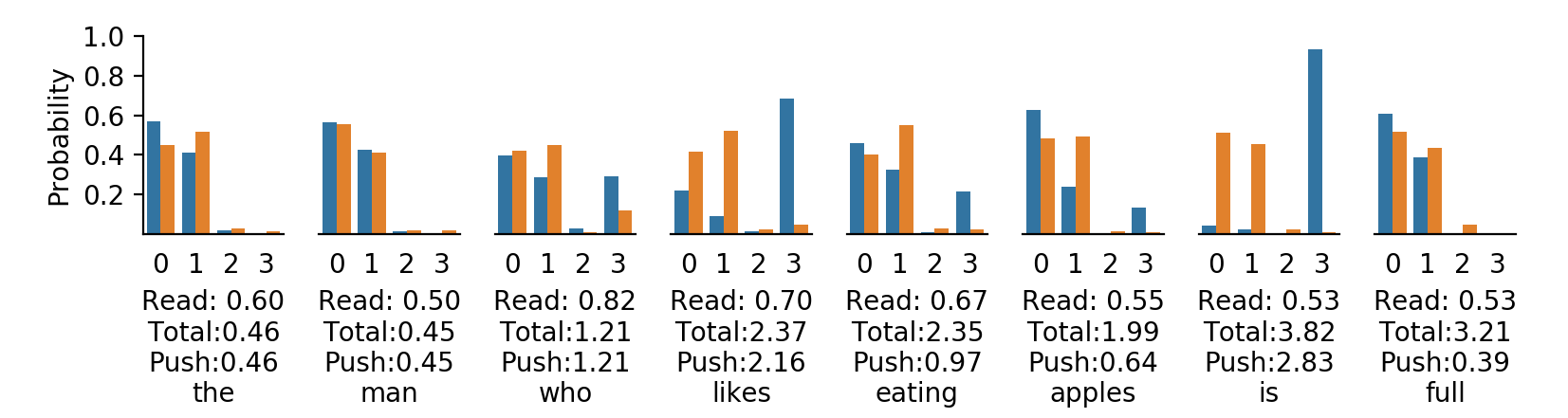} 
	     \caption{Distributions for push and read strengths at each step of processing example sentences. For example, the push strength chosen after processing \textit{the} ($0.46$) is the expected value of the blue distribution in the far left plot.}
	      \label{fig:sent1_dists}
\end{figure*}
\begin{figure*}[t]
	    \centering
	    \begin{tabular}{c c c}
	     \includegraphics[scale=0.3]{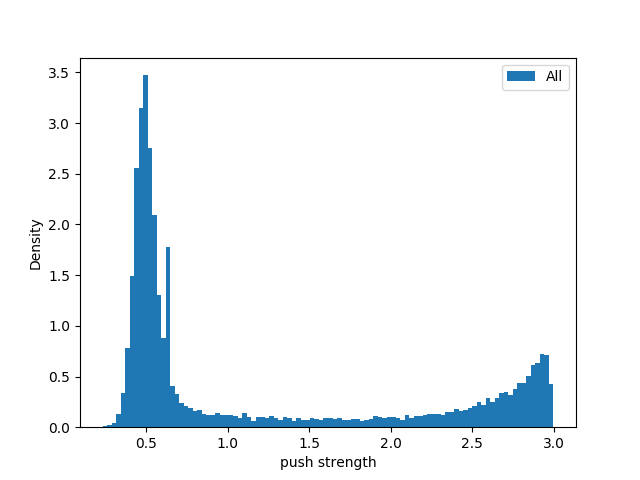} &
	     \includegraphics[scale=0.3]{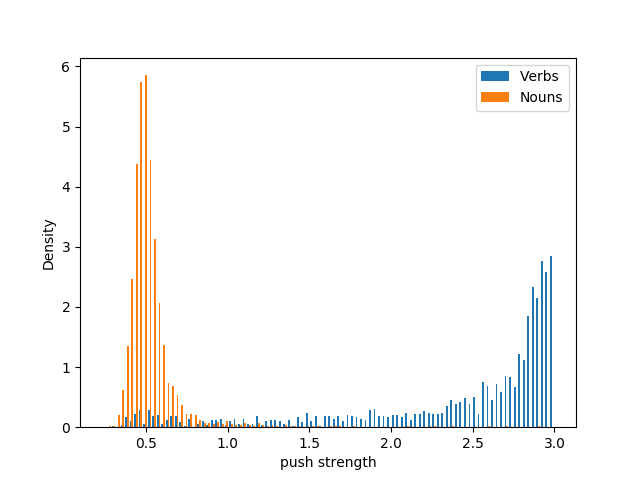} & 
	     \includegraphics[scale=0.3]{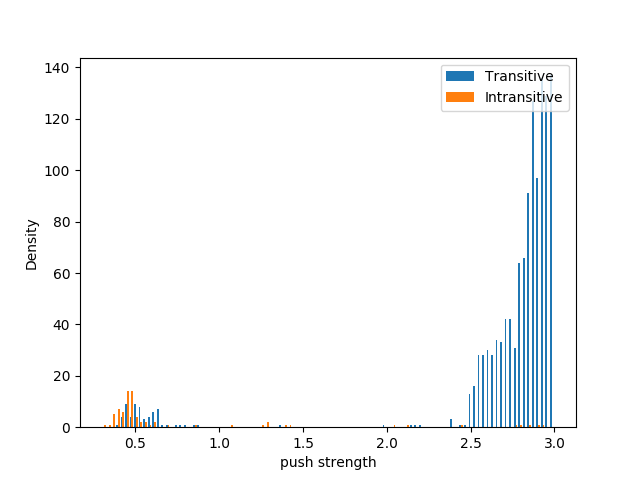} \\
	     All Words & Nouns and Verbs & Transitive vs. Intransitive Verbs
	    \end{tabular}
	    
	    \caption{Distributions of $d_t$ for the $u_t = 1$ language model over all test sentences. The center panel shows the distributions of $d_t$ for nouns and verbs, and the right panel shows the distributions for selected transitive and intransitive verbs.}
	    \label{fig:push_distributions}
\end{figure*}

The results presented in \autoref{sec:trainig_results} show that the $u_t = 1$ stack RNNs perform comparably to LSTMs in terms of quantitative evaluation metrics. The goal of this section is to assess whether or not stack RNNs achieve this level of performance in an interpretable manner. We do this by visualizing the push and read strengths computed by the $u_t = 1$ language model when processing two example sentences. These visualizations are shown in \autoref{fig:sent1_trace} and \autoref{fig:sent1_dists}. Notice that the push strength tends to spike following words with subcategorization requirements. For example, the preposition \textit{in} and the transitive verbs \textit{eat} and \textit{is} both require NP objects, and accordingly the model assigns a high push strength to these words. This suggests that the model is using the stack to capture hierarchical dependencies by keeping track of words that predictably introduce various kinds of phrases. 

\autoref{fig:push_distributions} shows push strengths computed by the $u_t = 1$ language model, aggregated across the entire Wikipedia corpus. We see that push strengths differ systematically based on part of speech. The distribution of push strengths computed by the network upon seeing a noun is tightly concentrated around $0.5$, whereas the push strength upon seeing a verb tends to be greater---usually more than $2.5$. This phenomenon reflects the fact that verbs typically take objects while nouns do not. 

We also find that push strengths assigned to verbs depend on their transitivity. The right panel of \autoref{fig:push_distributions} shows push strength distributions for a collection of common transitive and intransitive verbs. The model distinguishes between these two types of verbs by assigning high push strengths to transitive verbs and low push strengths to intransitive verbs. We make similar observations for other parts of speech: prepositions, which take objects, typically receive higher push strengths, while determiners and adjectives, which do not take phrasal complements, receive lower push strengths.

\section{Inference of Syntactic Structure} \label{sec:structure_inference}

\autoref{sec:interpreting_stack_usage} has shown that the push strengths $d_t$ computed by the $u_t = 1$ language model reflect the subcategorization requirements of the words encountered by the network. Based on this phenomenon, we may interpret the stack to be keeping track of phrases that are ``in progress.'' A high push strength induced by a transitive verb, for example, may be thought to indicate that a verb phrase has begun, and that this phrase ends when the object of the verb is seen. We thus hypothesize that for each time step $t$, $d_t$ represents the size of the phrase that begins with the word read by the network at time $t$. If $d_t$ is low, then this phrase consists of a single word; if it is high, then this is a complex phrase consisting of multiple words.

A similar intuition underlies the unsupervised parsing framework of \citet{shenNeuralLanguageModeling2018,shenStraightTreeConstituency2018}. Under this framework, constituency structure is induced from a sequence of words by computing a \textit{syntactic distance} between every two adjacent words. Intuitively, the syntactic distance between two words measures the distance from the lowest common parent node of the two words to the bottom of the tree. If two words have a low syntactic distance, then they are likely siblings in a small constituent; if they have a high syntactic distance, then they probably belong to different phrases. Whereas \autoref{fig:sent1_trace} and \autoref{fig:sent1_dists} allow us identify specific time steps at which the stack recognizes the beginning of a phrase, the unsupervised parsing framework allows us to explicitly visualize the phrasal organization of input sequences induced by our interpretation of the push strengths.

Given an input sequence $\mathbf{x}_1, \allowbreak \mathbf{x}_2, \allowbreak \dots, \allowbreak \mathbf{x}_n$, we define the syntactic distance between each $\mathbf{x}_t$ and $\mathbf{x}_{t - 1}$ for our $u_t = 1$ model to be the push strength $d_t$ computed by the controller during time $t$. If the current word does not open any new constituents, then it belongs to the same constituent as the previous word, and therefore should be assigned a low syntactic distance. On the other hand, if the current word opens a complex constituent, then it is lower in the parse tree than the previous word, and therefore should be assigned a high syntactic distance. 
Similarly, for our $d_t = 1$ model, we let $u_t$ be the syntactic distance between $\mathbf{x}_t$ and $\mathbf{x}_{t + 1}$. Under this interpretation, the pop strength estimates the complexity of the constituents that the current word closes. If the current word closes many complex constituents, then the next word appears at a higher level in the parse tree, and is therefore syntactically distant from the current word.  

\begin{algorithm}
	\caption{\citep{shenNeuralLanguageModeling2018,shenStraightTreeConstituency2018}}
	\label{alg:parsing}
	\begin{algorithmic}[1]
		\Procedure{MakeTree}{$\mathbf{X}, \mathbf{d}$} 
			\If{$\mathbf{X}$ has at most one word}
			    \State \Return $\mathbf{X}$
			\Else
			    \State $i \gets \argmax_j \mathbf{d}_j$
			    \State $l \gets \textsc{MakeTree}(\mathbf{X}[:i - 1], \mathbf{d}[:i - 1])$
			    \State $r \gets \textsc{MakeTree}(\mathbf{X}[i + 1:], \mathbf{d}[i + 1:])$
			    \If{$l$ and $r$ are not empty}
			        \State \Return $\tree[l, \tree[\mathbf{X}[i], r]]$
			    \ElsIf{$l$ is empty}
			        \State \Return $\tree[\mathbf{X}[i], r]$
			    \Else
			        \State \Return $\tree[l, \mathbf{X}[i]]$
			    \EndIf
			\EndIf
		\EndProcedure
	\end{algorithmic}
\end{algorithm}

\autoref{alg:parsing} shows our procedure for constructing trees. The algorithm takes as input a sequence of words arranged into a matrix $\mathbf{X}$ and a vector $\mathbf{d}$ containing the syntactic distance between each word and the previous word. Following \citet{shenNeuralLanguageModeling2018,shenStraightTreeConstituency2018}, we recursively split $\mathbf{X}$ into binary constituents. At each recursion level, we greedily choose the word with the highest syntactic distance as the split point. The final output is a binary tree spanning the full sentence.

\subsection{Evaluation}

We compute F1 scores for the parses obtained from our Wikipedia language models by comparing against parses from Section 23 of the Penn Treebank's Wall Street Journal corpus (WSJ23, \citealp{marcusPennTreebankAnnotating1994}). Since \autoref{alg:parsing} produces unlabeled binary trees, our evaluation uses the gold standard of \citet{htut-etal-2018-grammar-induction}, which consists of unlabeled, binarized versions of the WSJ23 trees. We also decapitalize the first word of every sentence for compatibility with our training data.

As a baseline, we the F1 scores attained by our models to those computed for purely right- and left-branching trees. A right-branching parse is equivalent to the output of \autoref{alg:parsing} on a sequence of equal syntactic distances. Thus, the difference between the right-branching F1 score and our models' scores is a measure of the amount of syntactic information encoded by the push and pop strength sequences. We also compare our F1 scores to the results of  \citeauthor{htut-etal-2018-grammar-induction}'s (\citeyear{htut-etal-2018-grammar-induction}) replication study for the \textit{parsing--reading--predict network} models (PRPN-LM and PRPN-UP), the two syntactic-distance-based unsupervised parsers originally proposed by \citet{shenNeuralLanguageModeling2018}.

\subsection{Results}

\begin{table}
    \small
	\centering
	
	\begin{tabular}{|lc|}
	    \hline
	    \textbf{Model} & \textbf{Parsing F1} \\
	    \hline
	    Stack ($u_t = 1$) & 31.2 \\
	    Stack ($d_t = 1$) & 16.0 \\
	    \hline
	    Right Branching & 13.1 \\
	    Left Branching & 7.3 \\
	    \hline
	    Best PRPN-UP \citep{htut-etal-2018-grammar-induction} & 26.3 \\
	    Best PRPN-LM \citep{htut-etal-2018-grammar-induction} & 37.4 \\
	    \hline
	\end{tabular}
	
	\caption{Unsupervised parsing performance evaluated on the WSJ23 dataset, attained by our stack models (top), the right- and left-branching baselines (middle), and the PRPN models (bottom).}
	\label{tab:parsing_results}
\end{table}

\begin{figure*}[h]
    \centering
    \begin{tabular}{c c}
    
        \begin{tikzpicture}[scale=0.5]
        \Tree [ [ The finger-pointing ] [ has [ [ already begun ] {.} ] ] ]
        \end{tikzpicture}
        &
        
        \begin{tikzpicture}[scale=0.5]
        \Tree [  [  The finger-pointing  ] [  [  [  has already  ] begun  ] {.} ]  ]
        \end{tikzpicture}
        \\
        
        \begin{tikzpicture}[scale=0.5]
        \Tree [ [ [ [ The futures ] halt ] [ was [ even assailed ] ] ] [ by [ [ Big [ Board [ floor traders ] ] ] {.} ] ] ]
        \end{tikzpicture} &
        
        \begin{tikzpicture}[scale=0.5]
        \Tree [  [  The [  futures halt  ]  ] [  [  [  was even  ] [  assailed [  by [  Big [  Board [  floor traders  ]  ]  ]  ]  ]  ] {.} ]  ]
        \end{tikzpicture}
        \\
        
        \begin{tikzpicture}[scale=0.5]
        \Tree [ [ CONCORDE [ trans-Atlantic flights ] ] [ are [ [ [ {\$} 2,400 ] [ to [ Paris [ and [ {\$} 3,200 ] ] ] ] ] [ to [ London {.} ] ] ] ] ]
        \end{tikzpicture} &
        
        \hspace{-.5em}\begin{tikzpicture}[scale=0.5]
        \Tree [  [  CONCORDE [  trans-Atlantic flights  ]  ] [  [  are [  [  [  [  {\$} 2,400  ] [  to Paris  ]  ] and  ] [  [  {\$} 3,200  ] [  to London  ]  ]  ]  ] {.}  ]  ]
        \end{tikzpicture}
    \end{tabular}
    \caption{Sample parses obtained from our stack RNN language model with $u_t = 1$ (left), compared to \citeauthor{htut-etal-2018-grammar-induction}'s (\citeyear{htut-etal-2018-grammar-induction}) gold-standard parses (right).}
    \label{fig:linzen_parse}
\end{figure*}
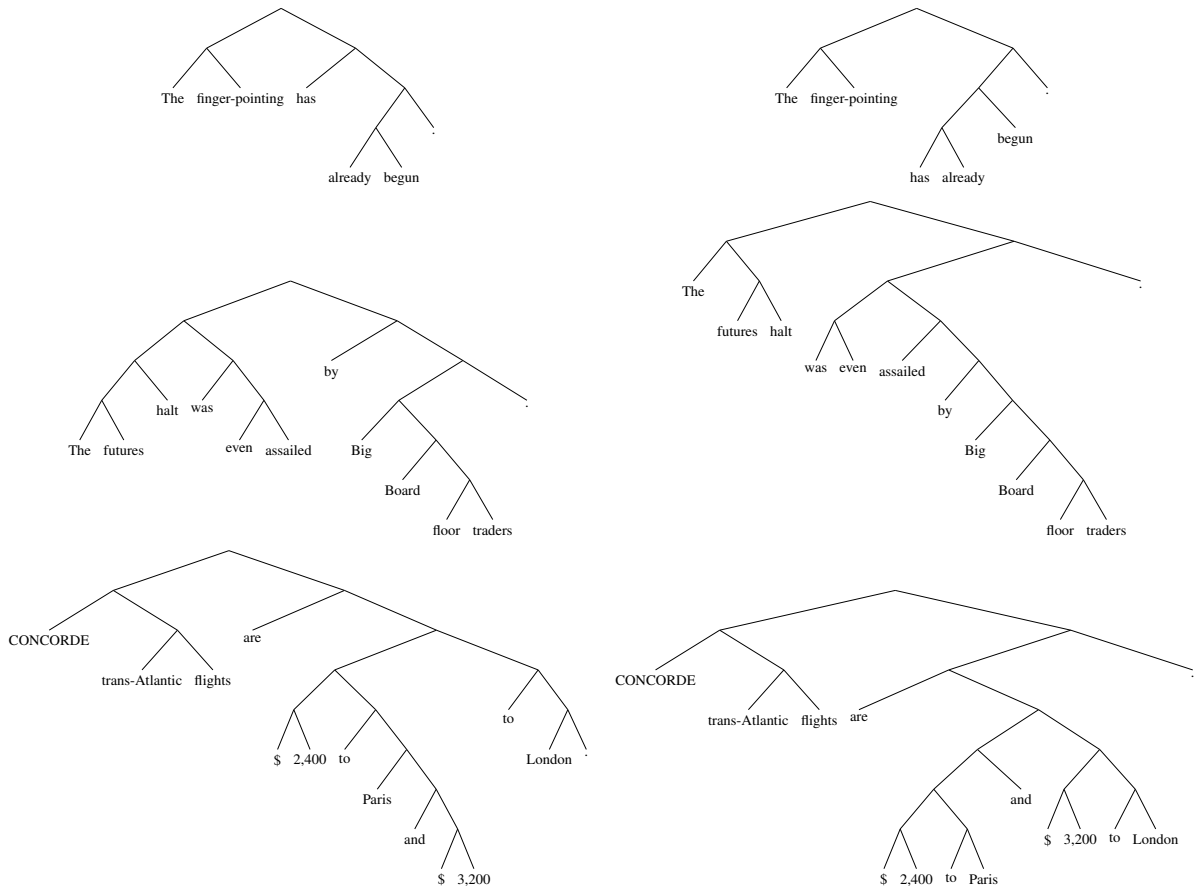

        
    
    

The F1 evaluation (see \autoref{tab:parsing_results}) shows that our Wikipedia model with $u_t = 1$ significantly outperforms the baseline on the Penn Treebank, while our model with $d_t = 1$ performs slightly better than the baseline. This is evidence that the types of hierarchical structures produced by \autoref{alg:parsing} resemble expert-annotated constituency parses.

Our results do not exceed those of \citeauthor{htut-etal-2018-grammar-induction}'s (\citeyear{htut-etal-2018-grammar-induction}) replication study. It is worth noting that our right- and left-branching baseline scores are somewhat lower than theirs. This suggests that differences in data processing or implementation might make our evaluation more difficult. Regardless, we consider our results to still be somewhat competitive, given that our language models were trained on out-of-domain data with few parameters and minimal hyperparameter tuning.

We provide example parses extracted from the stack RNN language models with $u_t = 1$ in \autoref{fig:linzen_parse}. 
Overall, our unsupervised parses tend to resemble the gold-standard parses with some differences. Periods in our parses systematically attach lower in the structure in our extracted parses than in the gold-standard trees.  High attachment would require a high syntactic distance (i.e., high push strength) between the period and the remainder of the sentence. However, the period inherently does not have any subcategorization requirements, so it induces a low push strength.  In contrast, prepositional phrases attach higher in our structures than in the gold parses. 
This may be the result of fixed subcategorization-associated push strengths for prepositions that give rise to fairly high estimates of syntactic distance. 

        

\section{Discussion} \label{sec:discussion}

Overall, our stack language models show no improvement over the LSTM baseline in terms of perplexity and classification accuracy. Although the $u_t = 1$ language model is comparable to the LSTM on these metrics, it ultimately achieves worse scores than the baseline. However, we have now seen that the pushing behavior of the model reflects subcategorization properties of lexical items that play an important role in determining their syntactic behavior, and that these properties allow reasonable parses to be extracted from this model. These observations show that the $u_t = 1$ model has learned to encode structural representations using the stack. Quantitatively, the importance of this structural information for the training objectives can be seen in \autoref{tab:classification_results}, where the stack at least partially alleviates the difficulty experienced by the LSTM classifier in handling syntactically complex inputs.

While our stack language models do not exceed the LSTM baseline in terms of perplexity and agreement accuracy, \citet{yogatamaMemoryArchitecturesRecurrent2018} find that their Multipop Adaptive Computation Stack architecture substantially outperforms a bare LSTM on these metrics. Compared to their models, we use fewer parameters and minimal hyperparameter tuning. Thus, it is possible that increasing the number of parameters in our controller may lead to similar increases in performance in addition to the structural interpretability that we have observed.

\section{Conclusion} \label{sec:conclusion}

The results reported here point to the conclusion that stack RNNs trained on corpora of natural language text do in fact learn to encode sentences in a hierarchically organized fashion.  We show that the sequence of stack operations used in the processing of a sentence lets us uncover a syntactic structure that matches standardly assigned structure reasonably well, even if the addition of the stack does not improve the stack RNN's performance over the LSTM baseline in terms of the language modeling objective.  We also find that using the stack RNN to predict the grammatical number of a verb results in better hierarchical generalizations in syntactically complex cases than is possible with stackless models. 
Taken together, these results suggest that the stack RNN model yields comparable performance to other architectures, while producing structural representations that are easier to interpret and that show signs of being linguistically natural.

\bibliography{acl2019}
\bibliographystyle{acl_natbib}

\appendix

\end{document}